\documentclass[letterpaper, 10 pt, conference]{ieeeconf}  

\IEEEoverridecommandlockouts                              

\makeatletter
\def\endfigure{\end@float}
\def\endtable{\end@float}
\makeatother

\usepackage{graphicx}
\usepackage{amsmath}
\usepackage{amssymb}
\usepackage[T1]{fontenc}
\usepackage{color}
\usepackage[normalem]{ulem}
\useunder{\uline}{\ul}{}

\usepackage{algorithm}
\usepackage{algpseudocode}
\usepackage{stfloats}
\usepackage{subfigure}
\usepackage{booktabs}
\usepackage{threeparttable}
\usepackage{multirow}
\usepackage{multicol}
\usepackage{makecell}
\usepackage{dcolumn}
\usepackage{color}

\newcommand{\eg}{e.g.,\ }
\newcommand{\Reffig}[1]{Figure~\ref{#1}}
\newcommand{\Refsec}[1]{Section~\ref{#1}}

\newcommand{\Refeq}[1]{Equation~\ref{#1}}
\newcommand{\Reftab}[1]{Table~\ref{#1}}

\usepackage{amsthm}

\title{\LARGE \bf
Multi-LVI-SAM: A Robust LiDAR-Visual-Inertial Odometry for Multiple Fisheye Cameras
}

\author{Xinyu Zhang$^{1}$, Kai Huang$^{1}$, Junqiao Zhao$^{*,2,3}$, Zihan Yuan$^{2,3}$, Tiantian Feng$^{1}$
\thanks{$^{1}$Xinyu Zhang, Kai Huang and Tiantian Feng are with the School of Surveying Geo-Informatics, Shanghai, China Tongji University, (e-mail: 2053869@tongji.edu.cn; huangkai@tongji.edu.cn; fengtiantian@tongji.edu.cn).}
\thanks{$^{2}$Junqiao Zhao and Zihan Yuan are with Department of Computer Science and Technology, School of Electronics and Information Engineering, Tongji University, Shanghai, China, and the MOE Key Lab of Embedded System and Service Computing, Tongji University, Shanghai, China (e-mail: zhaojunqiao@tongji.edu.cn; 2332062@tongji.edu.cn).}%
\thanks{$"{3}$Institute of Intelligent Vehicles, Tongji University, Shanghai, China}%
}

\UseRawInputEncoding

\begin{document}

\maketitle
\pagestyle{empty}
\thispagestyle{empty}

\begin{abstract}
 
We propose a multi-camera LiDAR-visual-inertial odometry framework, Multi-LVI-SAM, which fuses data from multiple fisheye cameras, LiDAR and inertial sensors for highly accurate and robust state estimation. 
To enable efficient and consistent integration of visual information from multiple fisheye cameras, we introduce a panoramic visual feature model that unifies multi-camera observations into a single representation. 
The panoramic model serves as a global geometric optimization framework that consolidates multi-view constraints, enabling seamless loop closure and global pose optimization, while simplifying system design by avoiding redundant handling of individual cameras.
To address the triangulation inconsistency caused by the misalignment between each camera's frame and the panoramic model's frame, we propose an extrinsic compensation method. 
This method improves feature consistency across views and significantly reduces triangulation and optimization errors, leading to more accurate pose estimation.
We integrate the panoramic visual feature model into a tightly coupled LiDAR-visual-inertial system based on a factor graph. 
Extensive experiments on public datasets demonstrate that the panoramic visual feature model enhances the quality and consistency of multi-camera constraints, resulting in higher accuracy and robustness than existing multi-camera LiDAR-visual-inertial systems.

\end{abstract}

\section{INTRODUCTION}

Simultaneous Localization and Mapping (SLAM) has been widely applied in autonomous driving, robotics, virtual reality, and indoor navigation.  
Although existing visual-inertial odometry (VIO) systems (\eg VINS-Mono \cite{VINS-Mono}, OpenVINS \cite{OpenVINS}) and LiDAR-inertial odometry (LIO) systems (\eg LIO-SAM \cite{LIO-SAM}, FAST-LIO2 \cite{FAST-LIO2}) have achieved significant progress, their performance in texture-less scenes or under aggressive motion is limited by the inherent weaknesses of individual sensors: visual systems suffer from scale ambiguity and narrow field-of-view (FoV), while LiDAR systems perform poorly in geometrically degraded environments.

To address these issues, LiDAR-visual-inertial odometry (LVIO) systems such as LVI-SAM \cite{LVI-SAM} and FAST-LIVO2 \cite{FAST-LIVO2} fuse multi-sensor data to enhance reliability.  
However, most existing LVIO systems utilize monocular cameras, which have limited FoV, leading to two major limitations: 
(1) inadequate environmental perception during rapid viewpoint changes, and  
(2) feature tracking failure in texture-less or repetitive scenes.  
These issues significantly compromise odometry reliability in complex environments.

To alleviate these limitations, multi-camera systems such as PAN-SLAM \cite{PAN-SLAM} and MCOV-SLAM \cite{MCOV} have been proposed.  
By capturing a wider FoV, they improve environmental perception and robustness.  
However, relying solely on visual sensors leads to failure in dark or feature-less environments.

To address these challenges, we propose Multi-LVI-SAM, a robust LVIO framework that integrates multiple fisheye cameras.  
Instead of independently fusing each fisheye view, which would significantly increase computation, we construct a panoramic visual feature model to unify multi-camera observations.  
Visual features from each camera are projected onto a normalized sphere, where the center of the sphere is the center of the panoramic model.
This model serves as a global geometric optimization framework that consolidates multi-view constraints, enabling seamless loop closure and global pose optimization while reducing system complexity.

Due to the offset between the fisheye camera center and the center of the panoramic model, triangulation based on the panoramic model can lead to inaccurate 3D point estimates.
To address this problem, we introduce an extrinsic compensation method based on rigorous geometric deduction.  
This method improves inter-view feature consistency and significantly reduces triangulation and optimization errors, resulting in more accurate long-term pose estimation.

Our main contributions are as follows:
\begin{itemize}
\item We propose a LiDAR-Visual-Inertial odometry system, {Multi-LVI-SAM}, which significantly expands environmental perception by mitigating the FoV limitations of monocular cameras, leading to highly accurate and robust pose estimation.

\item We introduce a {panoramic visual feature model} for multi-fisheye camera systems, which simplifies computation when visual sensors are coupled with IMU and LiDAR.

\item We propose an extrinsic compensation method to address the triangulation inconsistency caused by the misalignment between each camera's frame and the panoramic model's frame.

\item We conduct extensive experiments on public datasets (Newer College Dataset, M2DGR), demonstrating the accuracy, robustness, and effectiveness of {Multi-LVI-SAM}.
\end{itemize}

\section{RELATED WORK}

\subsection{Multiple Camera Visual Odometry}

Multi-camera systems enhance perception through wide FoV. 
PAN-SLAM \cite{PAN-SLAM} extends ORB-SLAM2 \cite{ORB-SLAM2} with fisheye cameras for panoramic coverage, but its spherical projection introduces computational overhead.
The triangulation process fails to account for the physical offset between camera centers and the virtual spherical center, leading to reconstruction inaccuracies.
MCOV-SLAM \cite{MCOV} proposes observability-based keyframe selection for omnidirectional loop closure, while MCVO \cite{MCVO} introduces adaptive feature weighting for heterogeneous cameras. 
Both estimate individual camera poses before body-frame alignment and do not construct a common global reference frame.
To address vision limitations, PIW-SLAM \cite{PIW} fuses fisheye cameras with IMU and wheel encoders. 
MCVIO \cite{MCVIO}, BAMF-SLAM \cite{BAMF}, MAVIS \cite{MAVIS}, and RMSC-VIO \cite{RMSC} all tightly couple data from multiple wide-FoV cameras with an IMU, jointly optimizing poses to enhance odometry tracking and localization accuracy. However, they compute the poses of each camera independently. 
\cite{Panoramic-LVO} combines panoramic cameras with LiDAR, but remains prone to motion drift.

\subsection{LiDAR-visual-inertial Odometry}

LiDAR, despite its high precision, depth perception capability, and robustness to lighting variations, faces challenges in geometrically degraded scenes. 
To address this, LiDAR-visual-(inertial) odometry systems have been proposed to leverage the complementary advantages of LiDAR, cameras, and IMU.
TVL-SLAM \cite{TVL-SLAM} associates visual and LiDAR features in a spherical coordinate system for accurate pose estimation. 
Suln-LIO \cite{Suln-LIO} integrates InEKF with surfel-based maps for precise odometry and flexible mapping. 
V-LOAM \cite{V-LOAM} uses visual-inertial odometry to initialize LiDAR mapping. 
Lvio-Fusion \cite{LVIO} fuses stereo cameras, LiDAR, IMU, and GPS via graph optimization and adaptive sensor weighting. 
CamVox \cite{CamVox} enables automatic LiDAR-camera calibration in target-free scenes.
LVI-SAM \cite{LVI-SAM} tightly couples LiDAR, visual, and IMU data on a factor graph for robust SLAM. 
R$^{2}$Live \cite{R2LIVE} and R$^{3}$Live \cite{R3LIVE} combine filter-based odometry with factor graph optimization. 
FAST-LIVO \cite{FAST-LIVO} and FAST-LIVO2 \cite{FAST-LIVO2} use sparse-direct alignment for efficient and accurate pose estimation. 
LiVisSfM \cite{LiVisSfM} jointly optimizes LiDAR and visual poses through incremental bundle adjustment. 
LIVER \cite{LIVER} employs deep learning to enhance robustness under varying lighting conditions. 
mVLINS \cite{mVLINS} decouples 6-DoF state estimation across LiDAR, visual, and IMU modules for improved adaptability.
However, most existing approaches rely on monocular cameras, which are inherently limited by their narrow field of view (FOV).

\section{LIDAR-VISUAL-INERTIAL ODOMETRY BASED ON MULTI-FISHEYE CAMERA}

\begin{figure*}[t]
    \centering
    \includegraphics[width=\textwidth]{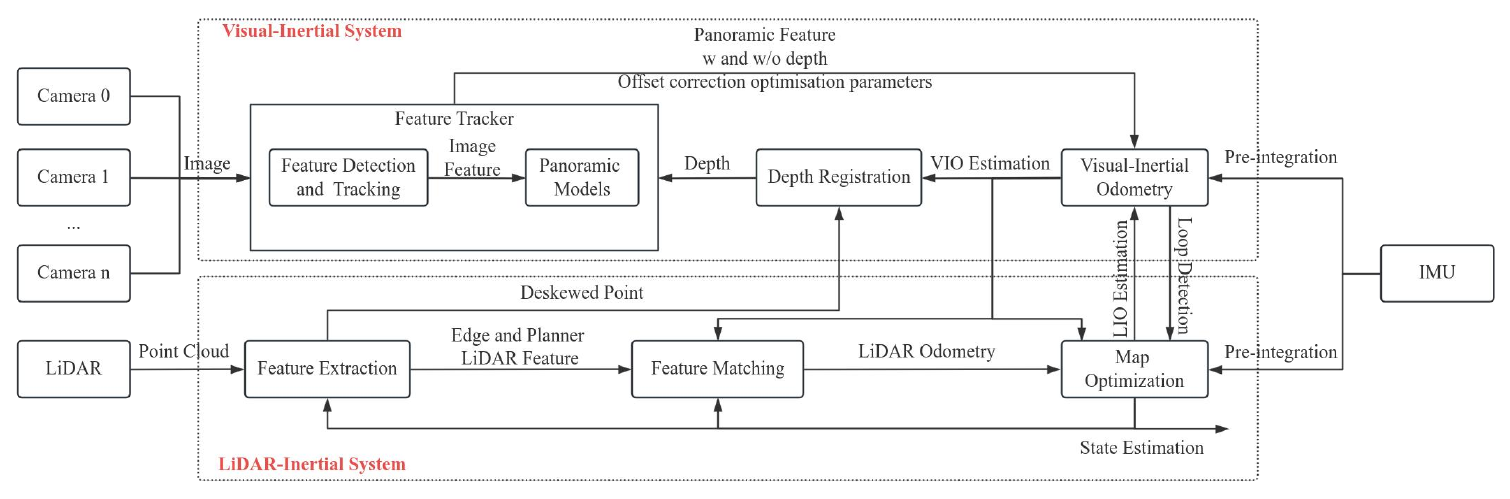}
    \caption{Overview of the proposed system.}
    \label{fig.system}
\end{figure*}

\subsection{System Overview} 

The proposed system (as shown in \Reffig{fig.system}) is built upon a tightly coupled dual-subsystem architecture to ensure robust state estimation, following the structure of LVI-SAM \cite{LVI-SAM}.

The VIO subsystem processes synchronized image and IMU data from multiple fisheye cameras to construct a unified panoramic visual feature model, which enables efficient fusion of multi-camera observations.  
It initializes the pose prior from LiDAR odometry and refines the state estimates by jointly minimizing the residuals of depth-enhanced visual features and IMU pre-integration.

The LIO subsystem extracts edge and planar features from LiDAR point clouds, performs scan-to-map matching for LiDAR odometry, and maintains a local feature map using a sliding window for real-time estimation.

Finally, IMU pre-integration constraints, visual odometry constraints, LiDAR odometry constraints, and loop closure constraints are jointly incorporated in a factor graph. The state estimation is solved in real time using iSAM2 \cite{iSAM2}.

\subsection{Visual-Inertial Odometry}

\begin{figure}[!hb]
    \centering
    \includegraphics[width=\columnwidth]{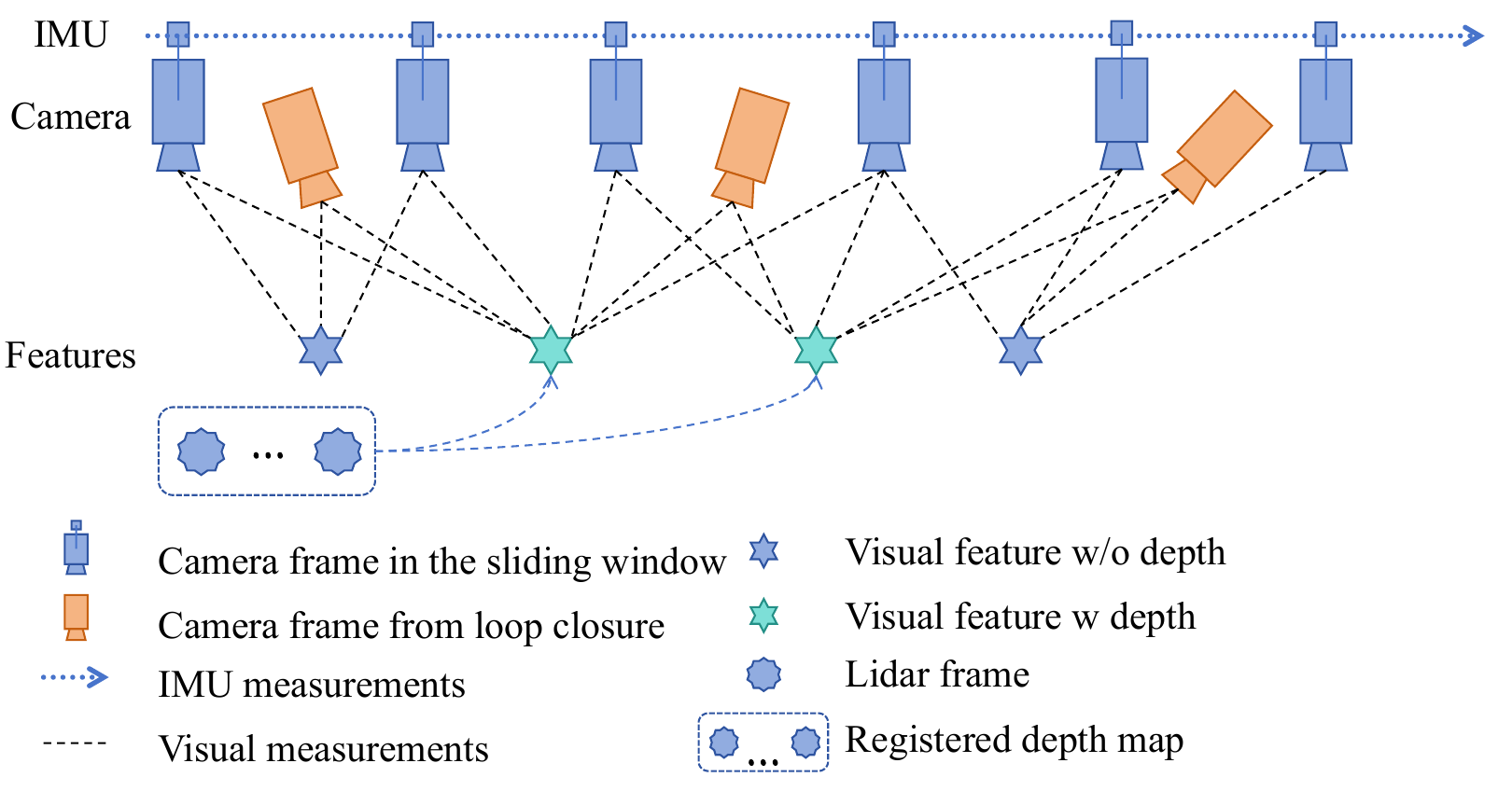}
    \caption{The framework of our visual-inertial odometry system, which initialises the attitude prior from LiDAR odometry and refines the estimates by minimising the joint residuals of the depth-enhanced visual features and the IMU pre-integration.}
    \label{fig.vio}
\end{figure}

The VIO subsystem is based on VINS-Mono \cite{VINS-Mono} (as shown in \Reffig{fig.vio}), with feature point extraction using the Shi-Tomasi algorithm \cite{Shi-Tomasi}, feature tracking using the Kanade-Lucas-Tomasi (KLT) optical flow tracking algorithm \cite{KLT}.
Multi-fisheye-cameras are integrated using a novel panoramic visual feature fusion model, which will be detailed in \Refsec{sec.pano}. 

Upon initialisation of the VIO subsystem, we register consecutive LiDAR point clouds by leveraging the initial pose estimate from visual odometry. 
This alignment helps generate a sparse depth image by projecting cumulated point cloud onto the camera's image plane.
This sparse depth data is then used to estimate depths for visual features.
The system then performs a sliding window non-linear optimization, where the states to be optimised are defined as follows:
\begin{equation}
    \begin{aligned} 
    \chi  &= \left[ {{{\bf{x}}_1}, \cdots ,{{\bf{x}}_n},{\bf{T}}_{\bf{c}}^{\bf{b}},{\lambda _{{d_1}}}, \cdots ,{\lambda _{{d_m}}}} \right]  \\
    {{\bf{x}}_k} &= \left[ {{\bf{p}}_{{b_k}}^w,{\bf{v}}_{{b_k}}^w,{\bf{q}}_{{b_k}}^w,{{\bf{b}}_a},{{\bf{b}}_g}} \right]\left( {k \in \left[ {0,n} \right]} \right) \\
    {\bf{T}}_c^b &= \left[ {{\bf{p}}_c^b,{\bf{q}}_c^b} \right]
    \end{aligned}
    \label{eq.statement}
\end{equation}
where ${{\bf{x}}_k}$ is the state vector of the IMU corresponding to the $k$th frame, ${{\bf{p}}_{{b_k}}^w}$ is the position of the IMU corresponding to the $k$th frame in the world coordinate system, ${{\bf{v}}_{{b_k}}^w}$ is the velocity of the IMU corresponding to the $k$th frame in the world coordinate system, ${{\bf{q}}_{{b_k}}^w}$ is the attitude of the IMU corresponding to the $k$th frame in the world coordinate system, ${{{\bf{b}}_a}}$ and ${{{\bf{b}}_g}}$ are the bias of accelerometer and gyroscope of the IMU, respectively, and $n$ is the total number of frames in the sliding window. 
${\bf{T}}_c^b$ is the extrinsic parameters between the camera and the IMU, ${\bf{p}}_c^b$ is the translation from the camera frame to the body frame, ${\bf{q}}_c^b$ is the quaternion representing the rotation from the camera frame to the body frame, ${\lambda_d}$ is the inverse depth of the 3D point in the coordinate system of the initial observation frame, and $m$ is the total number of feature points observed for the frames in the sliding window.


In following sections, all non-bold symbols represent geometric points, while bold symbols represent vectors that represent the coordinate of corresponding points. 


\subsubsection{Panoramic visual feature model} 
\label{sec.pano}

\begin{figure}[ht]
    \centering
    \includegraphics[scale=0.45]{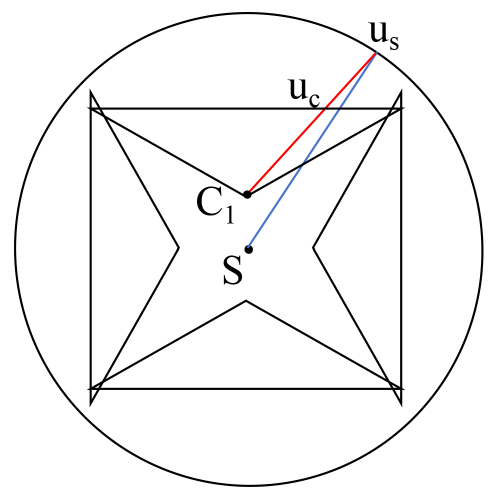}
    \caption{The panoramic visual feature model. Taking four cameras as an example, feature fusion is achieved by transforming each camera according to its extrinsic parameters with the panoramic visual feature model.}
    \label{fig.pano}
\end{figure}

To simplify system design by avoiding redundant handling of individual cameras, we propose a panoramic visual feature model for multi-fisheye cameras. 
As shown in \Reffig{fig.pano}, a spherical coordinate system is adopted in which $S$ is the center of the sphere, $C_1$ is the center of a camera, $u_c$ is a feature point in the fisheye camera coordinate system and $u_s$ is the feature point in the normalized sphere surface.

The transformation from $u_c$ to $u_s$ is as follows:
\begin{equation}
    \begin{aligned}
        \bf{u_s} &= \lambda {{\bf{R}}_i}{\bf{u_c}} + {{\bf{t}}_i} \\
        {\left\| {\bf{u_s}} \right\|^2} &= {r^2}
    \end{aligned}
    \label{eq.cam_to_pano}
\end{equation}
where ${{\bf{R}}_i}$ and ${{\bf{t}}_i}$ are the extrinsic parameters of the $i$th fisheye camera relative to the panoramic model, $r$ is the radius of the panoramic sphere, $\lambda$ is a scale to normalize the coordinate in the panoramic model.


\subsubsection{The extrinsic compensation for triangulation} 
\begin{figure}[ht]
    \centering
    \includegraphics[scale=0.45]{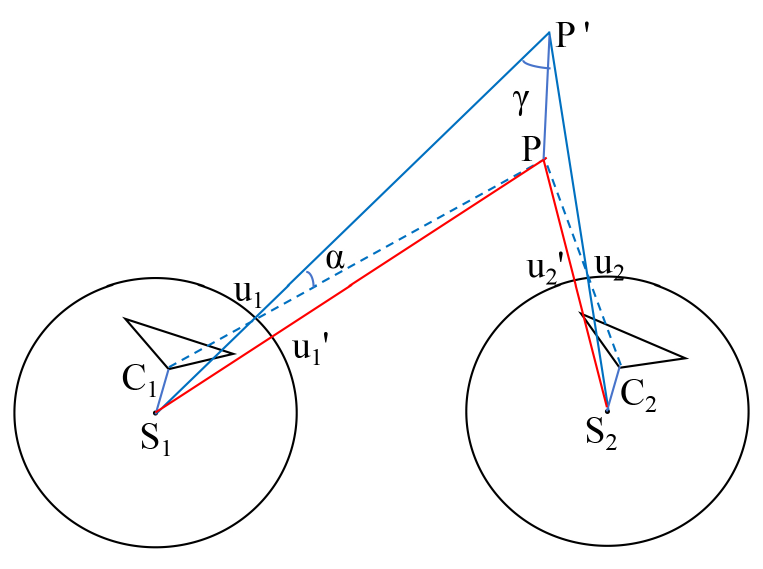}
    \caption{The existence of the translation offset between the center of the fisheye camera and the center of the panoramic sphere leads to an erroneous triangulation result, so we correct the triangulation error based on the extrinsic parameters between the center of the fisheye camera and the center of the panoramic sphere.}
    \label{fig.extrinsic-optimization}
\end{figure}

Although visual features from multiple cameras can be effectively represented in the panoramic model, triangulation can be problematic if the offset between a camera center and the center of the panoramic model is non-negligible.
As illustrated in \Reffig{fig.extrinsic-optimization}, the constraint required for accurate triangulation of point $P$ involves the camera centers ${C_1}$, ${C_2}$, and ${P}$'s corresponding projections ${u_1}$, ${u_2}$. 
However, due to the translational offset between the fisheye camera center and the panoramic model's center, triangulation based on the panoramic model uses the sphere centers $\S_1$, ${S_2}$, the projected points ${u_1}$, ${u_2}$, which results an incorrect point ${P'}$.
This leads to degraded state estimation performance.
 
To solve this problem, we propose an extrinsic compensation method for the deviation.

As shown in \Reffig{fig.extrinsic-optimization}, in the first frame, the camera center $C_1$, the projected points $u_1$ and the target point $P$ are collinear, and the sphere center $S_1$, the projected points $u_1$ and the error-triangulated point $P'$ are also collinear. 
Therefore, $C_1$, $S_1$, $u_1$, $P$ and $P'$ are coplanar. This holds similarly for other frames. 

Then, we define the normal of the plane formed by the camera center $C_1$, the panoramic sphere center $S_1$, and $u_1$ as $n_1$.
And the normal of the plane formed by $C_2$, $S_2$, and $u_2$ as $n_2$,
\begin{equation}
    {{\bf{n}}_1} = \overrightarrow {{S_1}{C_1}}  \times \overrightarrow {{S_1}{u_1}} , {{\bf{n}}_2} = \overrightarrow {{S_2}{C_2}}  \times \overrightarrow {{S_2}{u_2}} 
    \label{eq.2-optimization-normal-vector}
\end{equation}

Since we have the normal vectors of these two planes, we can compute the vector $\bf{m}$, which is parallel to the intersection vector $\overrightarrow {PP'}$ of these two planes. 
$\bf{m}$ can be derived as follows:  
\begin{equation}
    {\bf{m}} = {{\bf{n}}_1} \times {{\bf{n}}_2}
    \label{eq.2-optimization-parallel-vector}
\end{equation}

Then, the angle $\alpha$ between $\overrightarrow {{S_1}{u_1}}$ and $\overrightarrow {{C_1}{u_1}}$, and the angle $\gamma$ between $\overrightarrow {S_1P'}$ and $\overrightarrow {PP'}$ can be calculated as: 
\begin{equation}
    \begin{gathered}
        \alpha  = {\cos ^{ - 1}}(\frac{{\overrightarrow {{S_1}{u_1}}  \cdot (\overrightarrow {{S_1}{u_1}}  - \overrightarrow {{S_1}{C_1}} )}}{{\left\| {\overrightarrow {{S_1}{u_1}} } \right\| \cdot \left\| {\overrightarrow {{S_1}{u_1}}  - \overrightarrow {{S_1}{C_1}} } \right\|}}) \\
        \gamma  = {\cos ^{ - 1}}(\frac{{{\bf{m}} \cdot \overrightarrow {{S_1}{u_1}} }}{{\left\| {\bf{m}} \right\| \cdot \left\| {\overrightarrow {{S_1}{u_1}} } \right\|}})
    \end{gathered}
    \label{eq.2-optimization-angle}
\end{equation}

Let ${\lambda _{P'}}$ be the depth of $P'$ in the panoramic sphere coordinate before compensation, then we can derive the depth residual between $P$ and $P'$, ${\lambda _{PP'}}$ as: 
\begin{equation}
    {\lambda _{PP'}} = {\lambda _{P'}}\frac{{\sin (\alpha )}}{{\sin (\alpha  + \gamma )}}
    \label{eq.2-optimization-depth}
\end{equation}

So that,
\begin{equation}
    \bf{P} = {\bf{P'}} + {\lambda _{PP'}}{\bf{m}}
    \label{eq.2-optimization-result}
\end{equation}
where {\bf{P}} and {\bf{P'}} are the coordinates of points $P$ and $P'$, respectively.

When triangulating using multi-view, the 3D coordinate of a spatial point $X$ is determined by intersecting multiple planes, each constructed from the camera center $C_i$, the sphere center $S_i$ and the projected points $u_i$, as follows:

\begin{equation}
    \left[ {\begin{array}{*{20}{c}}
    {{{\bf{n}}_1}}\\
    {{{\bf{n}}_2}}\\
     \vdots \\
     {{{\bf{n}}_m}}
    \end{array}} \right] \cdot {\bf{\left[ {{\bf{R}}\left| {\bf{t}} \right.} \right]X}} = 0
    \label{eq.3-optimization}
\end{equation}
where ${\left[ {\begin{array}{*{20}{c}}
{{{\bf{n}}_1},}&{{{\bf{n}}_2},}&{ \cdots ,}&{{{\bf{n}}_m}}
\end{array}} \right]^ \top }$ are the normals of the plane formed by the camera center, the panoramic sphere center, and the feature point in the panoramic model from the same frame, calculated as in \Refeq{eq.2-optimization-normal-vector}. 
$\bf{\left[ {{\bf{R}}\left| {\bf{t}} \right.} \right]}$ is the extrinsic parameter between the world coordinate system and the panoramic coordinate system. 

\subsubsection{Feature depth association}
To improve the accuracy of VIO subsystem, we associate the depth from the point cloud with visual features.

\begin{figure}[ht]
    \begin{center}
        \subfigure[Depth association]{
            \centering
            \includegraphics[width=0.46\linewidth]{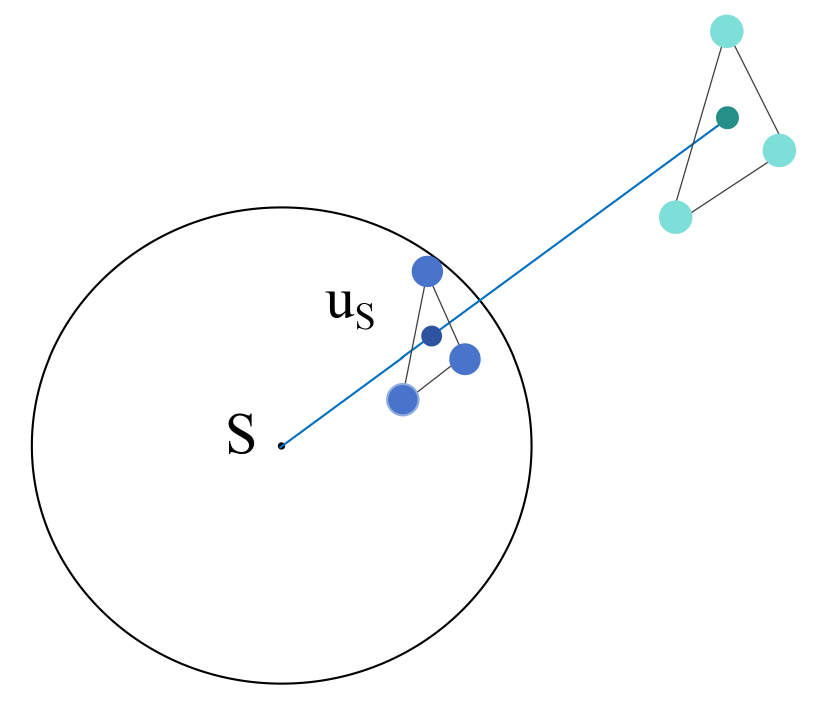}
            \label{fig.depth-association}
        }
        \subfigure[Occlusion induced depth blurring]{
            \centering
            \includegraphics[width=0.46\linewidth]{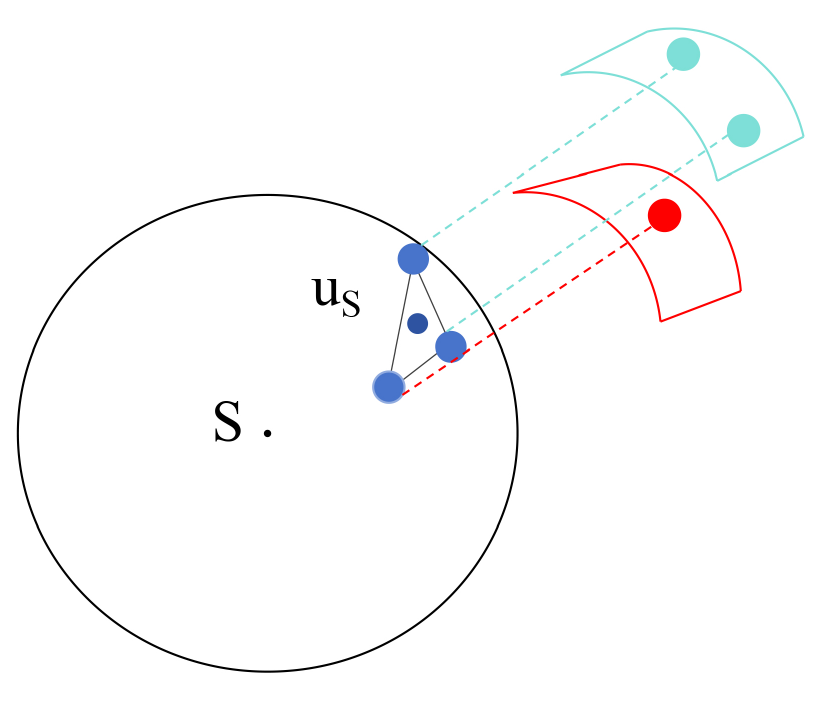} 
            \label{fig.blur-depth}
        }
        \caption{Feature depth association. In \subref{fig.depth-association} the cyan points represent the projected depth of point cloud and the blue point represent the visual feature on the normalized sphere. 
        In \subref{fig.blur-depth}, the cyan points and the red point represent the projected depth of point cloud, and the blue points represent the visual feature on the normalized sphere. 
        However, the cyan and red points come from lidar point clouds acquired from different frames, resulting in depth blurring.
        }
        \label{fig.depth}
    \end{center}
\end{figure}

Since a single LiDAR scan is sparse, we accumulate multi-frame LiDAR point clouds.
The accumulated point cloud is then projected onto the panoramic visual feature model. 
Similar to \cite{LVI-SAM}, we can estimate the depth of visual feature points by searching the nearest depth points on the panoramic sphere by conducting through a 2-D K-D tree (as illustrated in \Reffig{fig.depth}\subref{fig.depth-association}). 



As the depth map is obtained by accumulating multiple frames of the point cloud, depth blurring may occur due to the overlaying of objects (as shown in \Reffig{fig.depth}\subref{fig.blur-depth}).
Similar to \cite{LVI-SAM}, we avoid this problem by checking the maximum distance between the depth values of a feature.
If the distance exceeds a threshold, the feature is considered to have no reliable depth associated with it.

\begin{figure}[ht]
    \begin{center}
        \subfigure[camera0-visual]
        {
            \label{fig.camera0-visual}
            \includegraphics[width=0.45\linewidth]{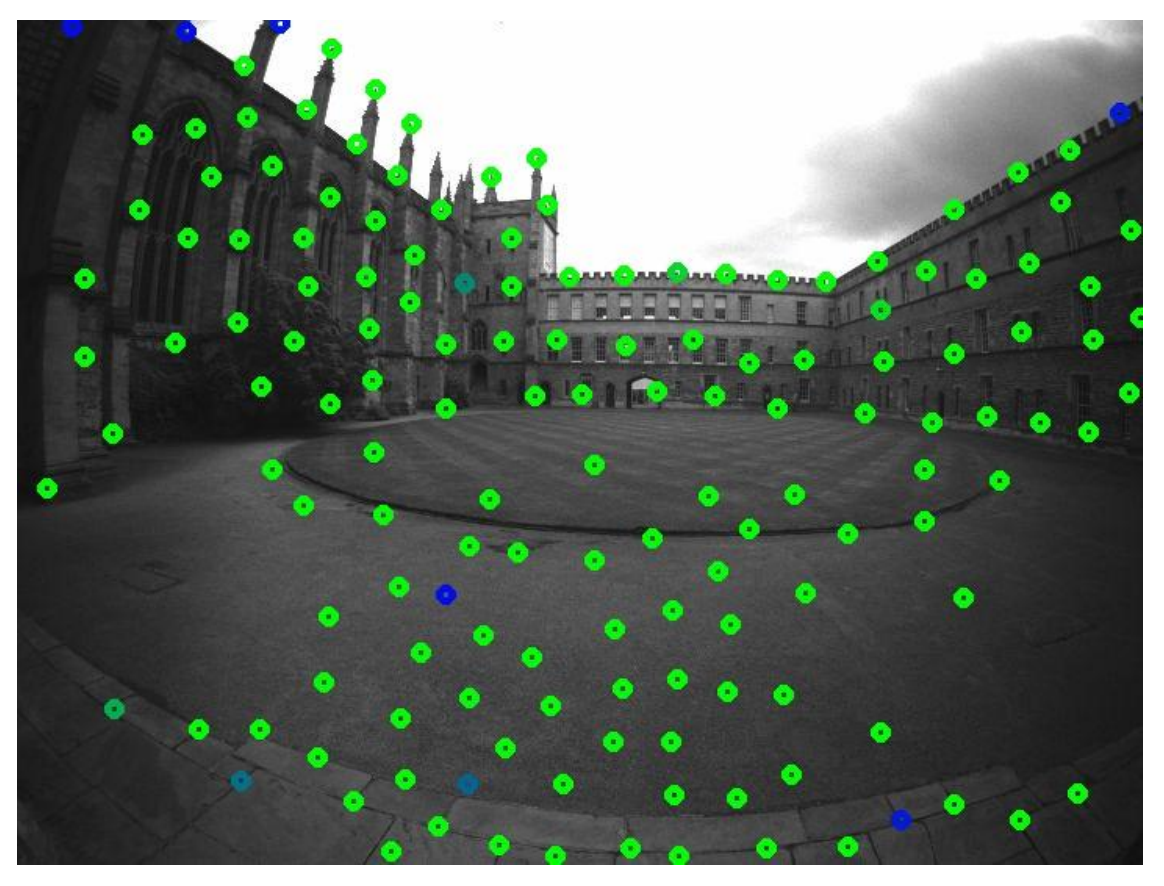}
        }
        \subfigure[camera0-depth]
        {
            \label{fig.camera0-depth}
            \includegraphics[width=0.45\linewidth]{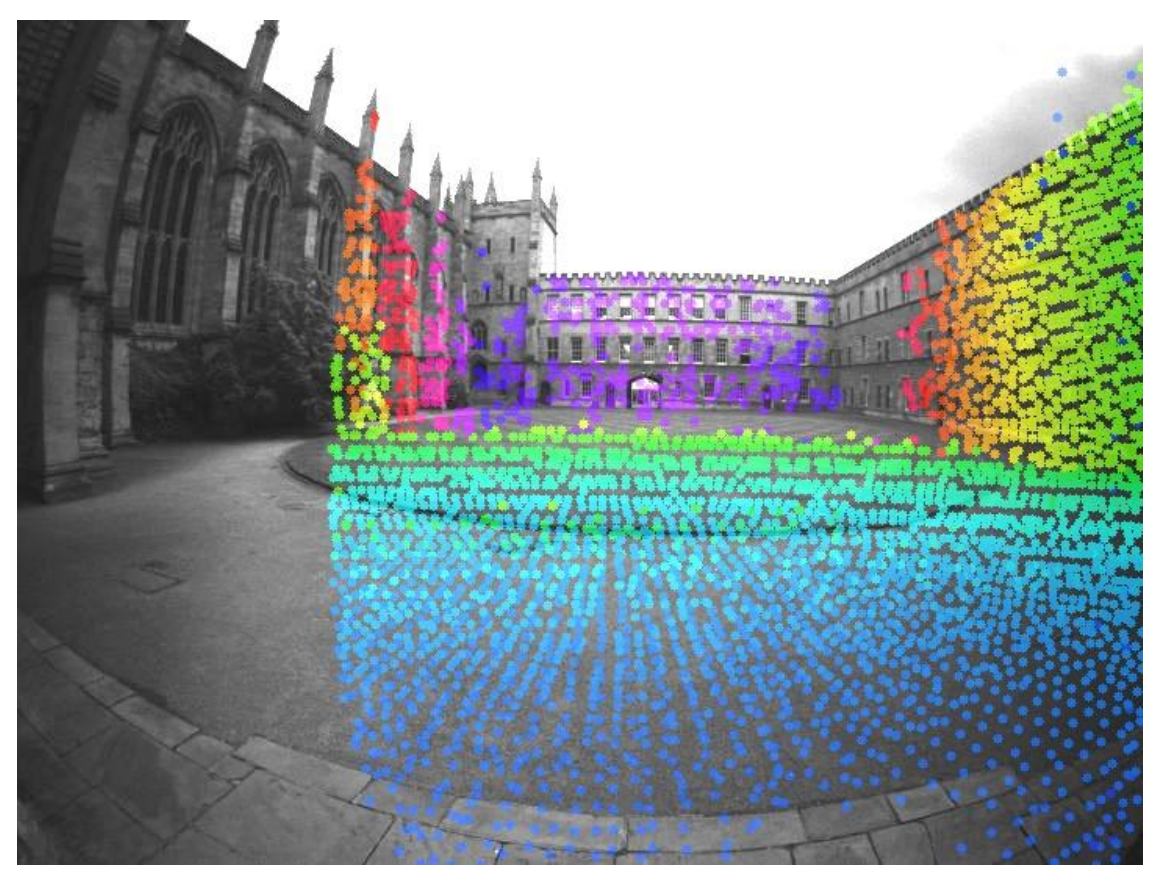}
        }
        \\
        \subfigure[camera1-visual]
        {
            \label{fig.camera1-visual}
            \includegraphics[width=0.45\linewidth]{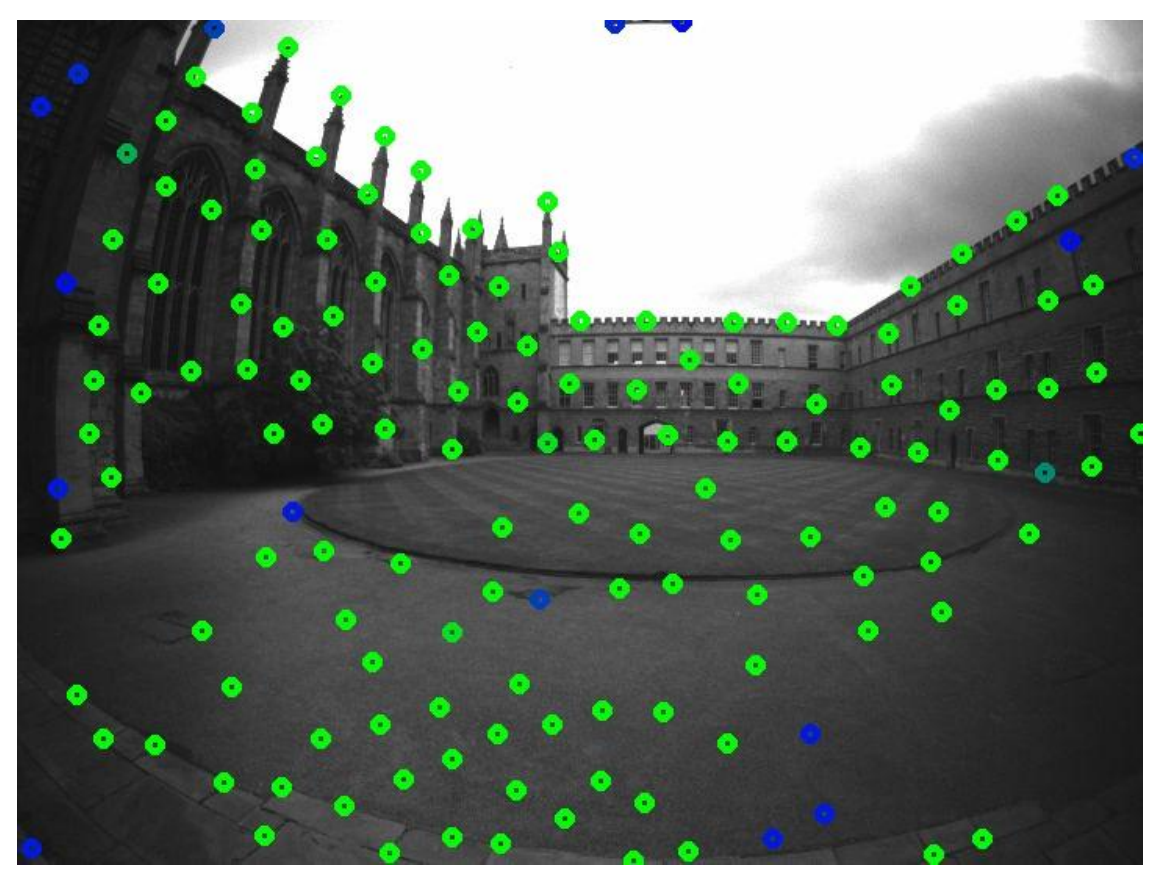}
        }
        \subfigure[camera1-depth]
        {
            \label{fig.camera1-depth}
            \includegraphics[width=0.45\linewidth]{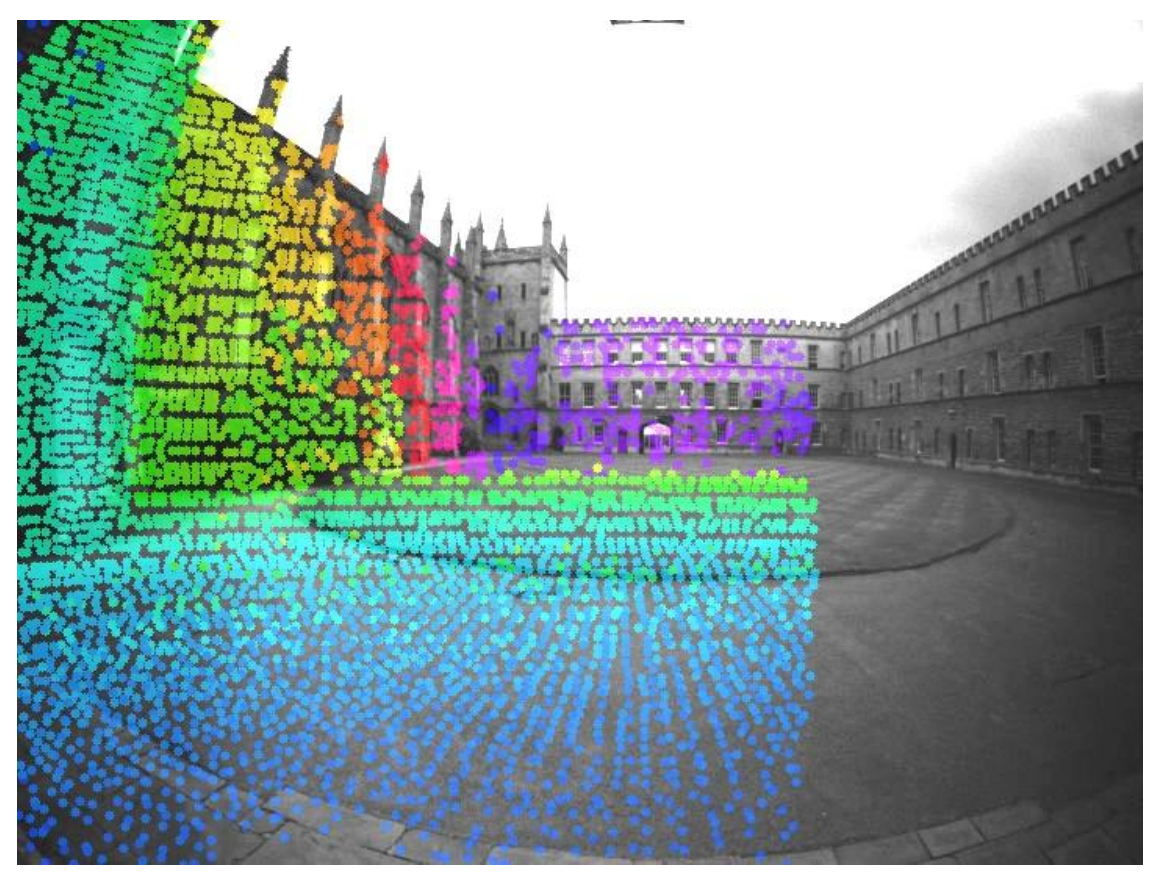}
        }
        \caption{The registered depth map and visual features.}
    \label{fig.vdmap}
    \end{center}
\end{figure}

A demonstration of the registered depth map and visual features is shown in \Reffig{fig.vdmap}. 
As shown in \Reffig{fig.vdmap} \subref{fig.camera0-visual}, \subref{fig.camera1-visual}, the green points indicate that the visual feature points are successfully associated with the depth information. 
The depth in the depth maps \subref{fig.camera0-depth}, \subref{fig.camera1-depth}, which are registered respective to the FoV of each camera, 
are projected onto the fisheye camera images. 
It is worth noting that the depth points in the overlapping areas of the fisheye cameras are projected onto the camera image whose FOV most likely contains the point cloud.

\subsubsection{Loop closure detection}
The DBoW2 \cite{DBoW2} algorithm is utilised for the loop detection. For each new keyframe of the VIO subsystem, BRIEF descriptors are extracted, and we fuse the information from multiple fisheye cameras using the panoramic visual feature model. Then we match them with the previously extracted descriptors. The loop closure candidate frames detected by DBoW2 are then sent to the LIO for further verification.
\subsection{LiDAR-Inertial Odometry}

\begin{figure}[!htbp]
    \centering
    \includegraphics[width=\linewidth]{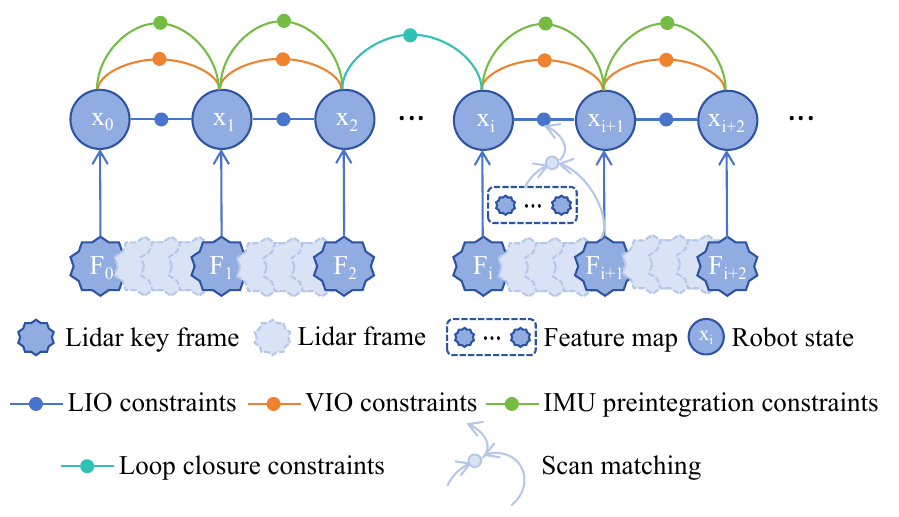}
    \caption{The framework of our LiDAR-inertial odometry system, which maintains a factor graph that has four types of constraints, including IMU preintegration constraints, VIO constraints, LIO constraints and loop closure constraints.}
    \label{fig.lio}
\end{figure}

The LIO subsystem is based on LIO-SAM \cite{LIO-SAM} and maintains a factor graph for global pose optimization (as shown in \Reffig{fig.lio}).
The IMU pre-integration constraints, visual odometry constraints, LiDAR odometry constraints, and loop closure constraints in the joint optimization factor graph are used to solve the state estimation problem.
LiDAR odometry constraints are derived from scan matching between LiDAR keyframes and the global feature map.
The loop closure constraints, initially proposed by the VIO subsystem, are further optimized by scan matching.
Furthermore, similar to visual odometry, we maintain the feature map through a sliding window for real-time estimation.

\section{Experiment}

\begin{table}[ht]
    \begin{center}
        \caption{Overview of the sensors}
        \label{tab.sensor}
        \resizebox{\linewidth}{!}{
        \begin{tabular}{p{25pt}<{\centering}p{30pt}<{\centering}p{55pt}p{20pt}<{\centering}p{90pt}}
            \hline\\[-2.9mm]\hline
            Dataset               & Sensor & Type            & Rate & Characteristics \\
            \hline
            \multirow{4}{25pt}{Newer College Dataset} & LiDAR  & Ouster & 10Hz & 128 Channels, 50 m Range \\
                                  & & OS0-128 & & $90^\circ$ Vertical FOV \\
                                  & Camera & Alphasense      & 30Hz & 720$\times$540\\
                                  & IMU    & Bosch BMI085    & 200Hz & 6-axis \\
            \hline
            \multirow{5}{25pt}{M2DGR Dataset} & LiDAR  & Velodyne & 10Hz & 32 Channels \\ 
                                  & & VLP-32C & & $40^\circ$ Vertical FOV \\
                                  & Camera & FLIR Pointgrey & 15Hz & 1280$\times$1024\\
                                  & & CM3-U3-13Y3C & & \\
                                  & IMU    & Handsfree A9 & 150Hz & 9-axis \\
            \hline\\[-2.9mm]\hline
        \end{tabular}
        }
    \end{center}
\end{table}

The proposed method has been evaluated on two public datasets: the Newer College Dataset \cite{newcollege} and the M2DGR Dataset \cite{m2dgr}, with sensor parameters detailed in \Reftab{tab.sensor}. State-of-the-art open source methods were used for the comparison experiments, including FAST-LIO2 \cite{FAST-LIO2}, LVI-SAM \cite{LVI-SAM}, FAST-LIVO2 \cite{FAST-LIVO2} and R$^{2}$Live \cite{R2LIVE}. 
All algorithms are implemented in C++ and tested on a computer with Intel i7-9800X CPU and Ubuntu OS.

\subsection{Newer College Dataset}

\begin{table*}[ht]
    \begin{center}
        \caption{Absolute translational errors (RMSE, meters) in Newer College Dataset}
        \label{tab.aten}
        \begin{tabular}{lcccccc}
            \hline\\[-2.9mm]\hline
            Sequence & FAST-LIO2 & Ours-LIO & FAST-LIVO2 & LVI-SAM (w/ loop) & Multi-LVI-SAM (w/o loop) & Ours (w/ loop) \\
            \hline
            Quad-Easy & 0.070734 & 0.072749 & \underline{0.070027} & fail & 0.071064 & \textbf{0.068595} \\
            \hline
            Quad-Hard & \underline{0.080453} & fail & \textbf{0.070212} & fail & 0.089902 & 0.089189 \\
            \hline
            Math-Easy & 0.111047 & 0.080668 & 0.131684 & fail & \underline{0.080243} & \textbf{0.080021} \\
            \hline
            Math-Medium & 0.118592 & fail & 0.128942 & fail & \underline{0.110313} & \textbf{0.107045} \\
            \hline
            Math-Hard & \textbf{0.067132} & fail & 0.103742 & fail & 0.089415 & \underline{0.88219} \\
            \hline
            Undermine-Hard & \textbf{0.057463} & fail & 0.149832 & fail & 0.077561 & \underline{0.077231} \\
            \hline
            Stairs & fail & 3.032322 & fail & fail & 0.701162 & \textbf{0.451100} \\
            \hline
            Cloister & 0.092785 & fail & 0.277620 & fail & \underline{0.081100} & \textbf{0.080777} \\
            \hline\\[-2.9mm]\hline
        \end{tabular}
    \end{center}
\end{table*}

The Newer College Dataset is collected using a handheld multi-camera LiDAR inertial system. 
The evaluation process involves comparison in both outdoor and indoor sequences, covering challenging scenarios such as textureless walls, rapid viewpoint changes, and severe motion. 
The benchmark is evaluated by using the root-mean-square error (RMSE) of absolute trajectory error. 

As shown in \Reftab{tab.aten}, the proposed method demonstrates superior performance in most sequences. 
In test sequences with relatively smooth sensor motion (such as the Easy series, Stairs, and Cloister), the proposed method demonstrates optimal performance in both localization accuracy and system stability, thanks to the richer visual observation data provided by the multi-camera system. 
Specifically, the multi-camera configuration not only effectively mitigates the inherent feature sparsity issue of monocular vision but also significantly improves depth estimation accuracy through our triangulation compensation method, enabling the system to maintain stable pose estimation even under varying lighting conditions or in low-texture environments.  

In the challenging Stairs scenario, both FAST-LIO2 and FAST-LIVO2 experience localization failures: FAST-LIO2 diverges due to point cloud degradation in narrow spaces, while FAST-LIVO2 is limited by the field of view and feature extraction capability of a monocular camera, unable to obtain sufficient visual constraints on low-texture stair surfaces.   
In contrast, the proposed method, through the collaborative observation of the multi-camera system, maintains stable feature tracking and pose estimation even in geometrically repetitive and texture-scarce environments like stairwells. 
The redundant perspectives provided by the multi-camera system not only expand the effective observation range but also reduce the probability of mismatches through multi-view geometric constraints, thereby enhancing overall system robustness while ensuring accuracy.  
This advantage is also validated in scenarios with uneven feature distribution, such as long corridors (Cloister), where the multi-camera system consistently provides reliable visual observations, avoiding performance degradation issues that single-sensor systems face in specific environments.

However, in sequences with rapid orientation changes or severe device shaking (Hard series), the performance of our LiDAR subsystem degrades considerably. 
Since the visual subsystem relies on accurate LiDAR point clouds to provide depth information for visual feature points, this adversely affects the overall system accuracy. 
But FAST-LIO2, which is also the LIO subsystem of FAST-LIVO2, performs registration based on planar features and the point-to-plane data association provides more effective constraints for pose estimation than the point-to-line data association of the LIO subsystem of the proposed method.

LVI-SAM employs a normalized plane in its backend optimization, which fails to fully leverage the wide FOV of fisheye cameras in the dataset. 
Furthermore, severe distortion in the peripheral regions of fisheye images often leads to misalignment with point clouds during visual-lidar data fusion. 
The proposed method uses panoramic visual feature model which significantly enhances feature richness address these limitations.

\subsection{M2DGR Dataset}

\begin{table*}[ht]
    \begin{center}
        \caption{Absolute translational errors (RMSE, meters) in M2DGR Dataset}
        \label{tab.atem}
        \begin{tabular}{lccccccc}
            \hline\\[-2.9mm]\hline
            Sequence & FAST-LIO2 & Ours-LIO & FAST-LIVO2 & R$^{2}$Live & LVI-SAM (w/ loop) & Ours (w/o loop) & Ours (w/ loop) \\
            \hline
            room-01 & 0.314317 & \underline{0.135437} & fail & fail & fail  & 0.138986 & \textbf{0.135218} \\
            \hline
            room-02 & 0.314317l & \underline{0.127656} & fail & fail & fail & 0.134496 & \textbf{0.127308} \\
            \hline
            room-03 & 0.412898 & \underline{0.161249} & 0.172044 & 0.271782 & fail & 0.210872 & \textbf{0.161111} \\
            \hline
            hall-05 & 1.177671 & 1.029866 & \textbf{0.870324} & 1.371839 & fail & 1.030859 & \underline{0.992787} \\
            \hline
            door-02 & 0.28419 & 0.194134 & \textbf{0.157198} & 0.207294 & fail & 0.193460 & \underline{0.190772} \\
            \hline
            gate-03 & 0.186100 & \underline{0.107744} & 0.324915 & 0.514011 & fail & 0.108942 & \textbf{0.107078} \\
            \hline
            walk-01 & 0.110538 & 0.076908 & 0.148344 & 0.207999 & fail & \underline{0.076424} & \textbf{0.076371} \\
            \hline\\[-2.9mm]\hline
        \end{tabular}
    \end{center}
\end{table*}

\begin{table*}[ht]
    \begin{center}
        \caption{Absolute translational errors (RMSE, meters) in sequences}
        \label{tab.ateas}
        \resizebox{\textwidth}{!}{
        \begin{tabular}{llccccccc}
            \hline\\[-2.9mm]\hline
            \multirow{2}{20pt}{Dataset} & \multirow{2}{20pt}{Sequence} & camera0/ & camera1/ & camera2/ & camera3/ & Ours & Ours & Ours \\
             & & camera-left & camera-right & camera-midleft & camera-midright & (w/o camera) & (w/o compensation) & (w/ compensation) \\ 
            \hline
            \multirow{4}{25pt}{Newer College Dataset} & Quad-Hard & 0.091785 & 0.090536 & 0.089477 & 0.089629 & fail & \underline{0.089224} & \textbf{0.089189} \\
            \cline{2-9}
            & Math-Hard & 0.089151 & 0.089138 & 0.090447 & 0.089855 & fail & \underline{0.088492} & \textbf{0.088219} \\
            \cline{2-9}
            & Stairs & 0.676187 & \underline{0.468999} & 0.602967 & 0.574191 & 3.032322 & 0.759511 & \textbf{0.451100} \\
            \cline{2-9}
            & Cloister & \underline{0.086457} & 0.087435 & 0.088665 & 0.089718 & fail & 0.088408 & \textbf{0.080777} \\
            \hline
            \multirow{4}{20pt}{M2DGR Dataset} & room-01 & 0.139807 & 0.229812 & 0.136615 & 0.136706 & 0.138433 & \underline{0.135437} & \textbf{0.135218} \\
            \cline{2-9}
            & hall-05 & 1.100235 & 1.363355 & 1.173142 & \underline{1.021673} & 1.029866 & 1.068760 & \textbf{0.992787} \\
            \cline{2-9}
            & door-02 & 0.192287 & 0.193666 & 0.192226 & 0.193145 & 0.194134 & \underline{0.191782} & \textbf{0.190772} \\
            \cline{2-9}
            & walk-01 & 0.076897 & 0.076795 & \underline{0.076557} & fail & 0.076908 & 0.079851 & \textbf{0.076371} \\
            \hline
            \multirow{4}{20pt}{Hilti'2022 Dataset} & Construction Upper Level 3 & fail & fail & fail & 0.550198 & fail & \underline{0.171185} & \textbf{0.165085} \\
            \cline{2-9}
            & Basement 2 & 0.692945 & \underline{0.178062} & 0.294170 & 0.242493 & fail & 0.187401 & \textbf{0.166250} \\
            \cline{2-9}
            & Attic to Upper Gallery 2 & 0.892412 & \underline{0.868133} & 1.094798 & 1.442255 & 3.918399 & 0.917128 & \textbf{0.655462} \\
            \cline{2-9}
            & Corridor Lower Gallery 2 & 0.839022 & 0.977379 & 1.084816 & fail & fail & \underline{0.689255} & \textbf{0.517772} \\
            \hline\\[-2.9mm]\hline
        \end{tabular}
        }
    \end{center}
\end{table*}

M2DGR Dataset is collected by a ground platform, which includes indoor, outdoor and long-trajectory sequences.
As shown in \Reftab{tab.atem}, the proposed method demonstrates superior performance in most sequences.
It is noteworthy that in the sequences room-01 and room-02, FAST-LIVO2 and R$^{2}$Live fail to estimate accurately and their trajectories drift substantially.
This primarily stems from two factors: First, the presence of dynamic objects (such as moving pedestrians) in the scene disrupts the continuity of visual features, making it difficult for monocular or single-camera systems to maintain stable inter-frame feature matching. 
Second, during prolonged operation, the inherent scale uncertainty of monocular systems leads to continuously accumulating errors, ultimately causing the entire system to diverge. 
In contrast, the proposed multi-camera system leverages the multi-view observations to effectively filter out interference features caused by dynamic objects, while the redundancy of multi-source observations significantly enhances the robustness of feature tracking. 

However, in the sequences such as hall-05 and door-02, the presence of large glass wall interferes with the depth information from LiDAR point clouds, which in turn affects the depth association of visual features and reduces the system's accuracy.  
LVI-SAM suffers from the same issue as in the Newer College Dataset.

\subsection{Ablation Study}
We conduct ablation experiments to validate the effectiveness of multi-camera data fusion and the proposed extrinsic compensation approach: (1) the use of a single camera; (2) the disabling of all cameras (LIO); and (3) with or without extrinsic compensation. 
The experiments are conducted on the Newer College Dataset, the M2DGR Dataset. 
And to further validate the robustness of the multi-camera system, we select the Hilti'2022 Dataset \cite{Hilti} for additional ablation experiments, which is collected using handheld devices and covers indoor and outdoor sequences in environments such as construction sites, galleries, exhibition halls, and basements. 
These sequences present numerous challenges, including long corridors, staircases, textureless features, poor lighting, and insufficient LiDAR planar constraints. and the Hilti'2022 Dataset. The results are shown in \Reftab{tab.ateas}.


\begin{figure}[ht]
    \begin{center}
        \subfigure[camera0-walk]
        {
            \label{fig.camera0-walk}
            \includegraphics[width=0.45\linewidth]{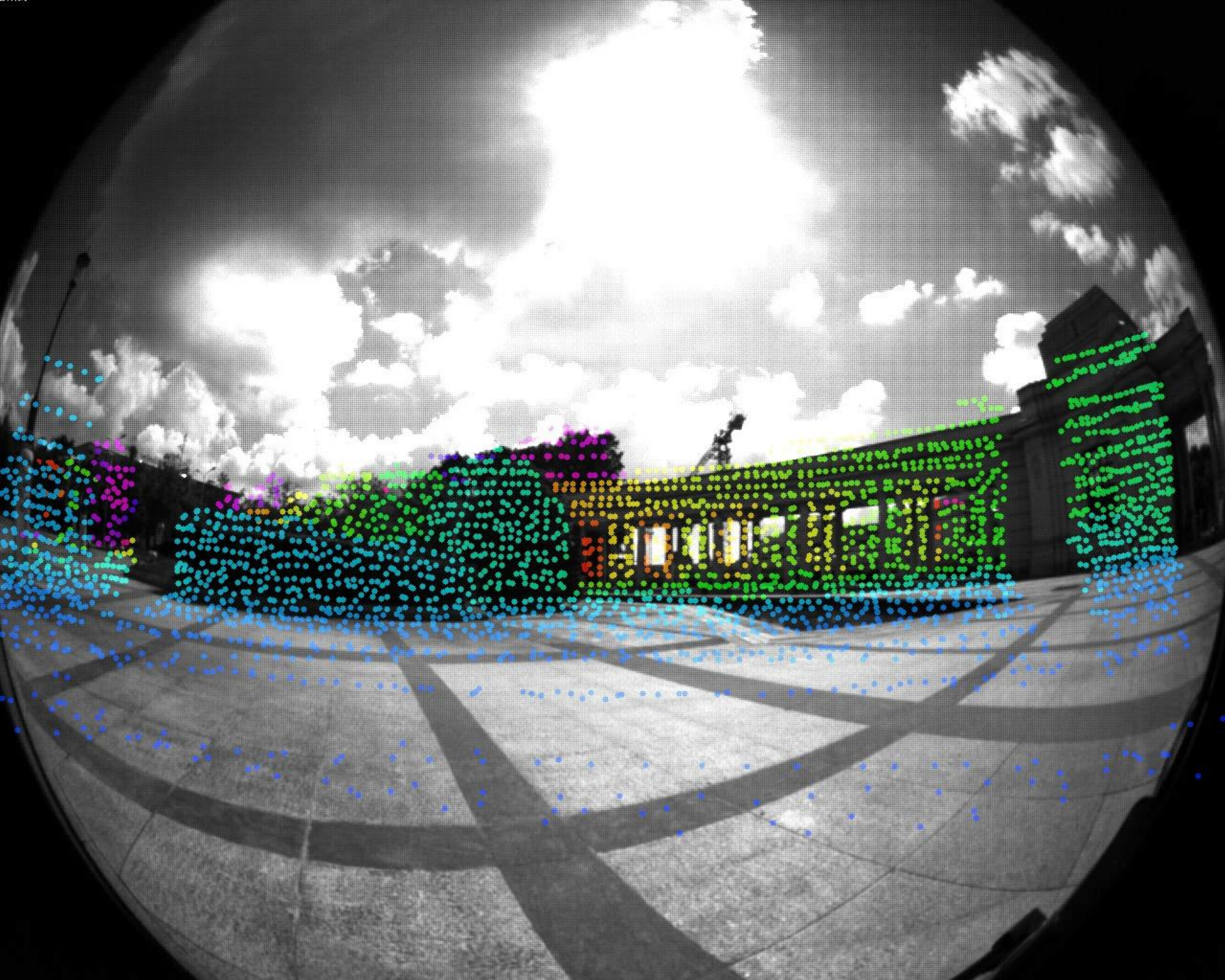}
        }
        \subfigure[camera1-walk]
        {
            \label{fig.camera1-walk}
            \includegraphics[width=0.45\linewidth]{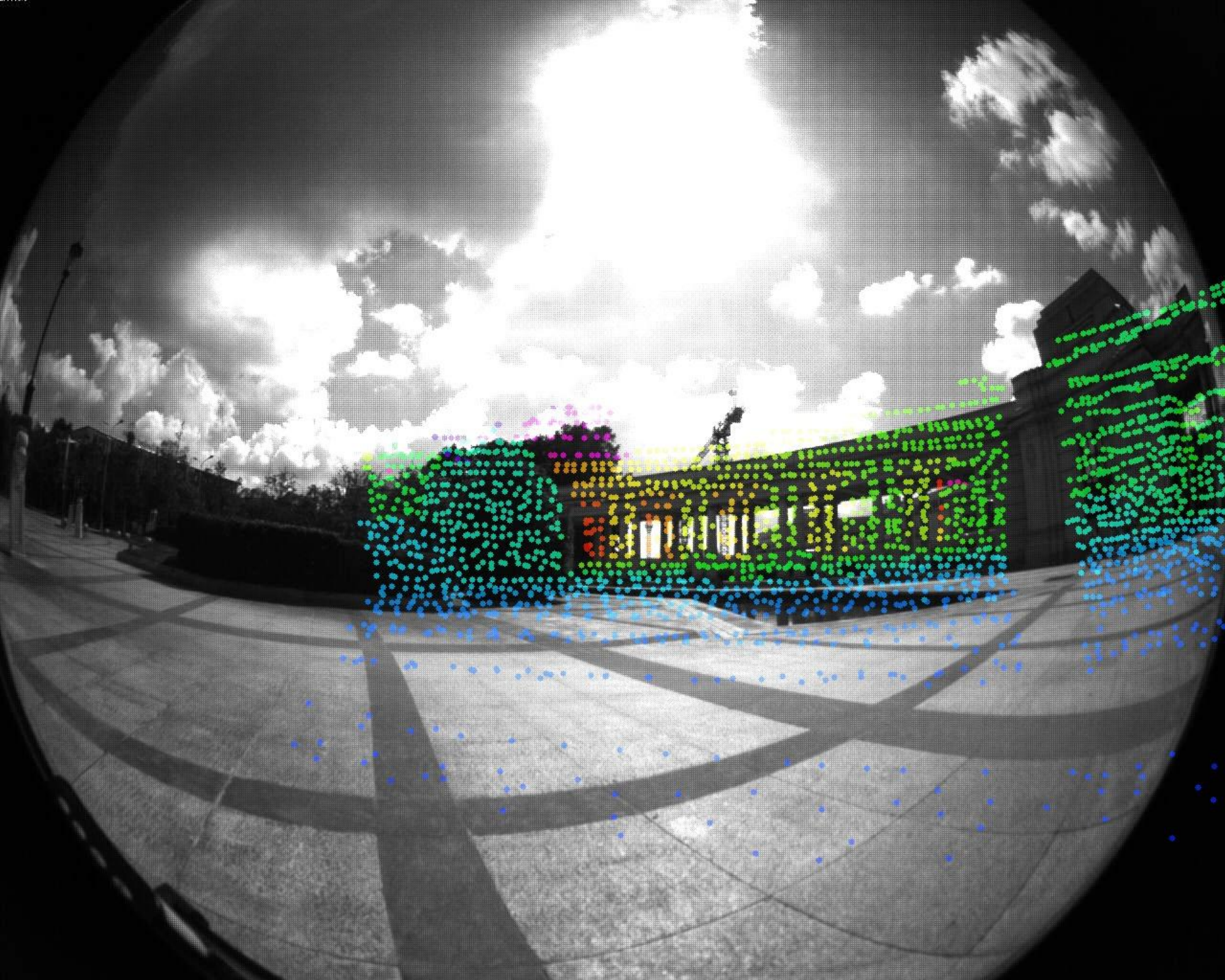}
        }
        \\
        \subfigure[camera2-walk]
        {
            \label{fig.camera2-walk}
            \includegraphics[width=0.45\linewidth]{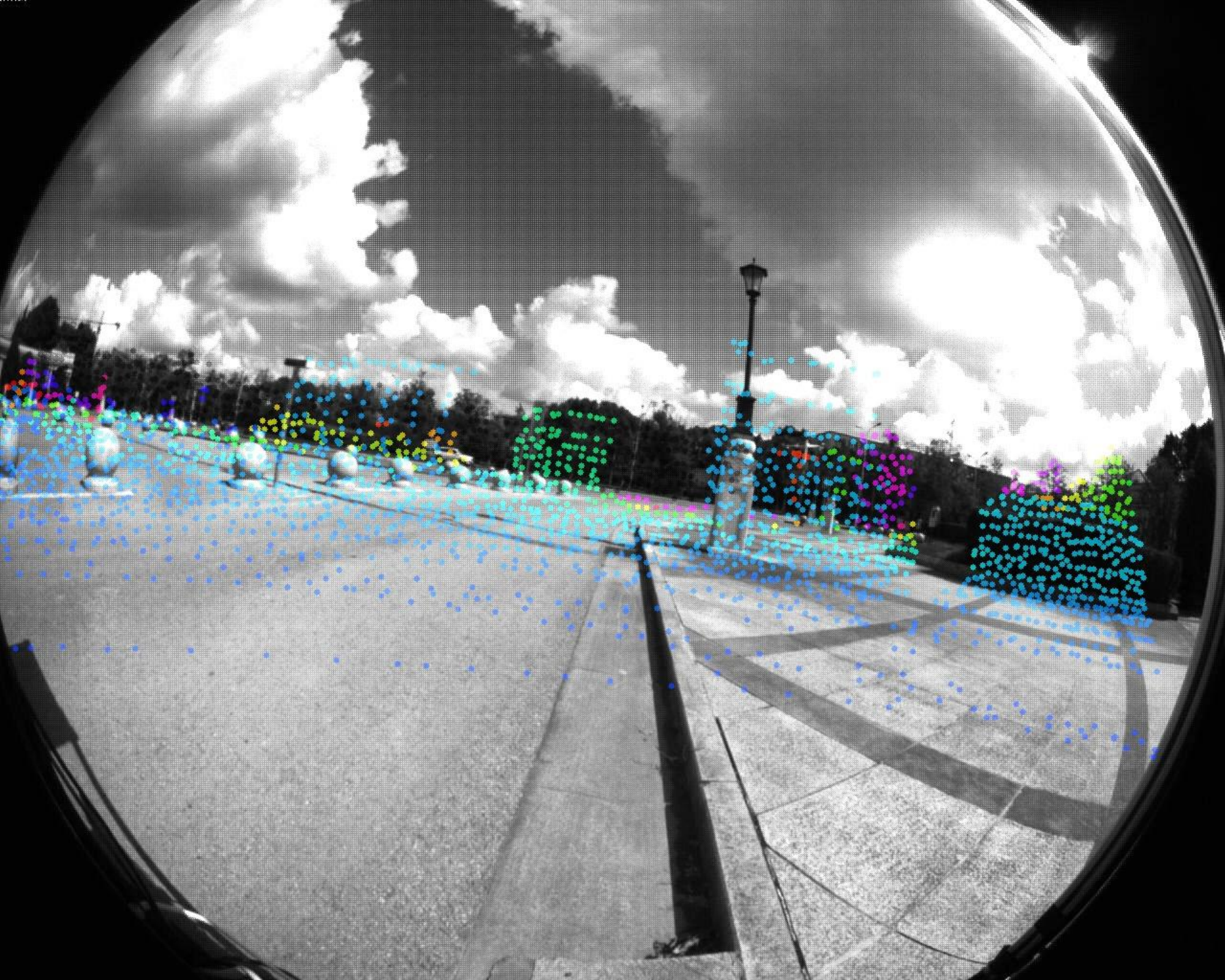}
        }
        \subfigure[camera3-walk]
        {
            \label{fig.camera3-walk}
            \includegraphics[width=0.45\linewidth]{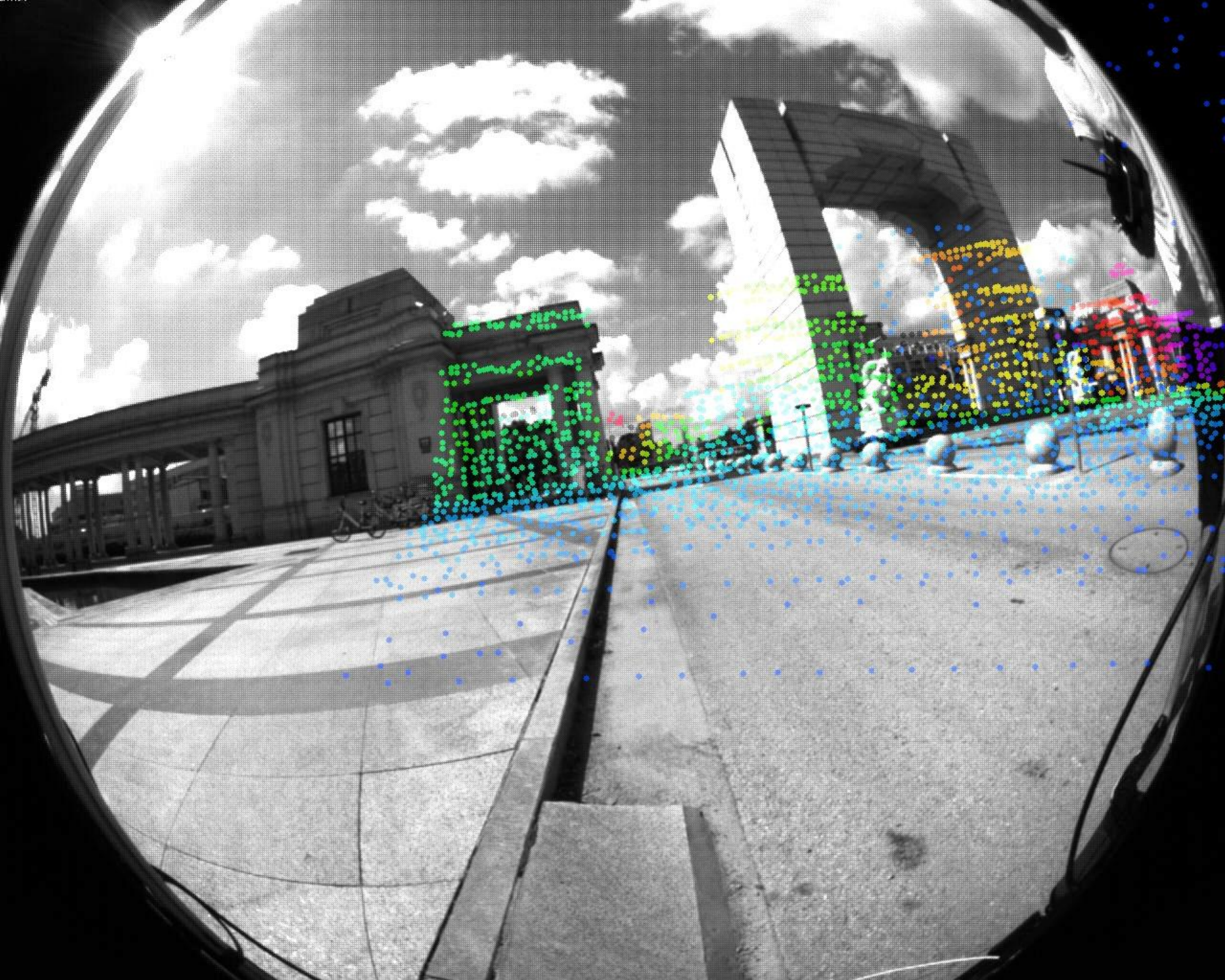}
        }
        \caption{The registered depth map in the walk-01 sequence.}
    \label{fig.walk-figure}
    \end{center}
\end{figure}

\begin{figure}[ht]
    \begin{center}
        \subfigure[Construction Upper Level 3]{
            \centering
            \includegraphics[width=0.48\linewidth]{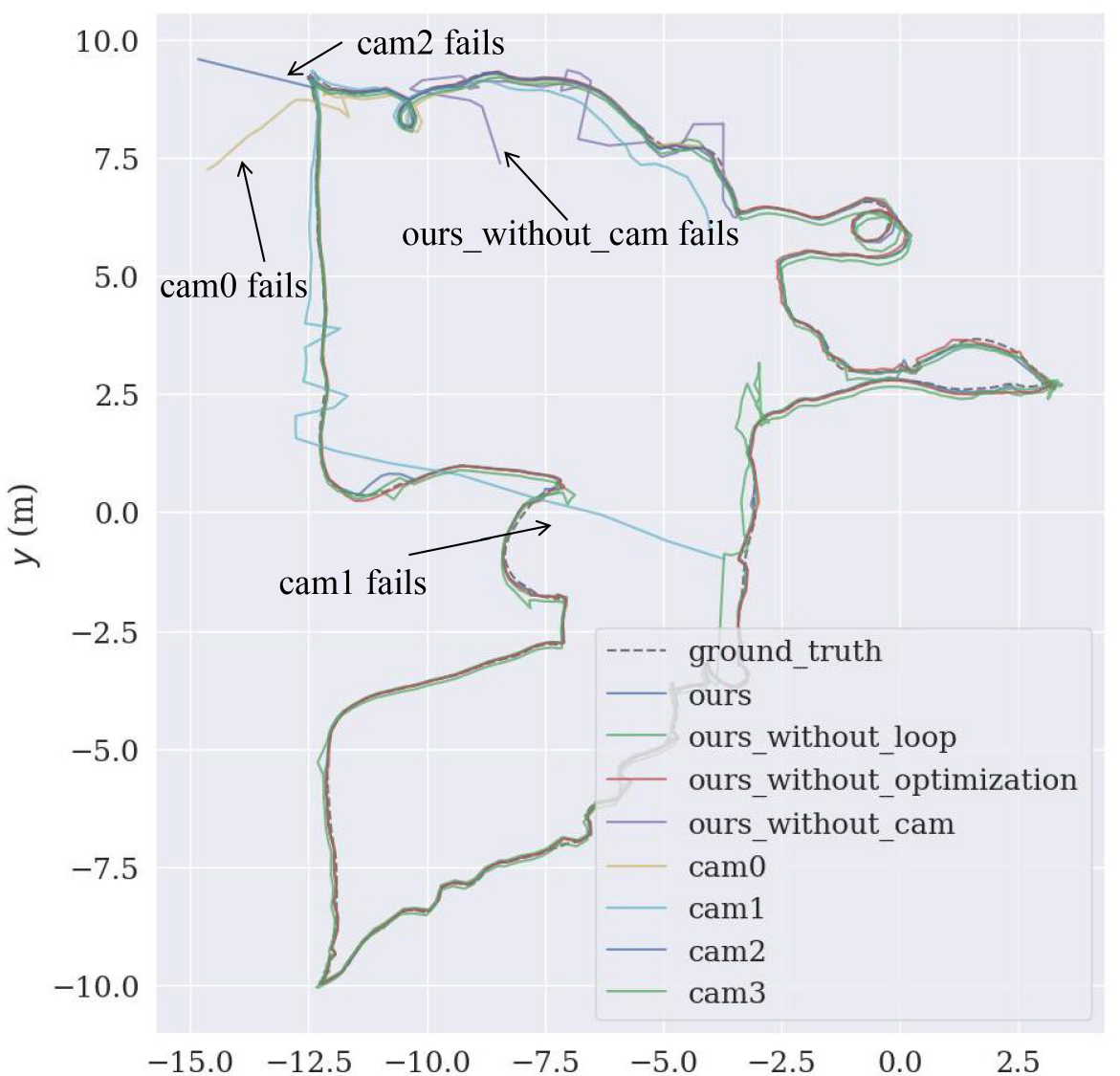}
            \label{fig.Construction-Upper-Level-3}
        }
        \subfigure[Corridor Lower Gallery 2]{
            \centering
            \includegraphics[width=0.45\linewidth]{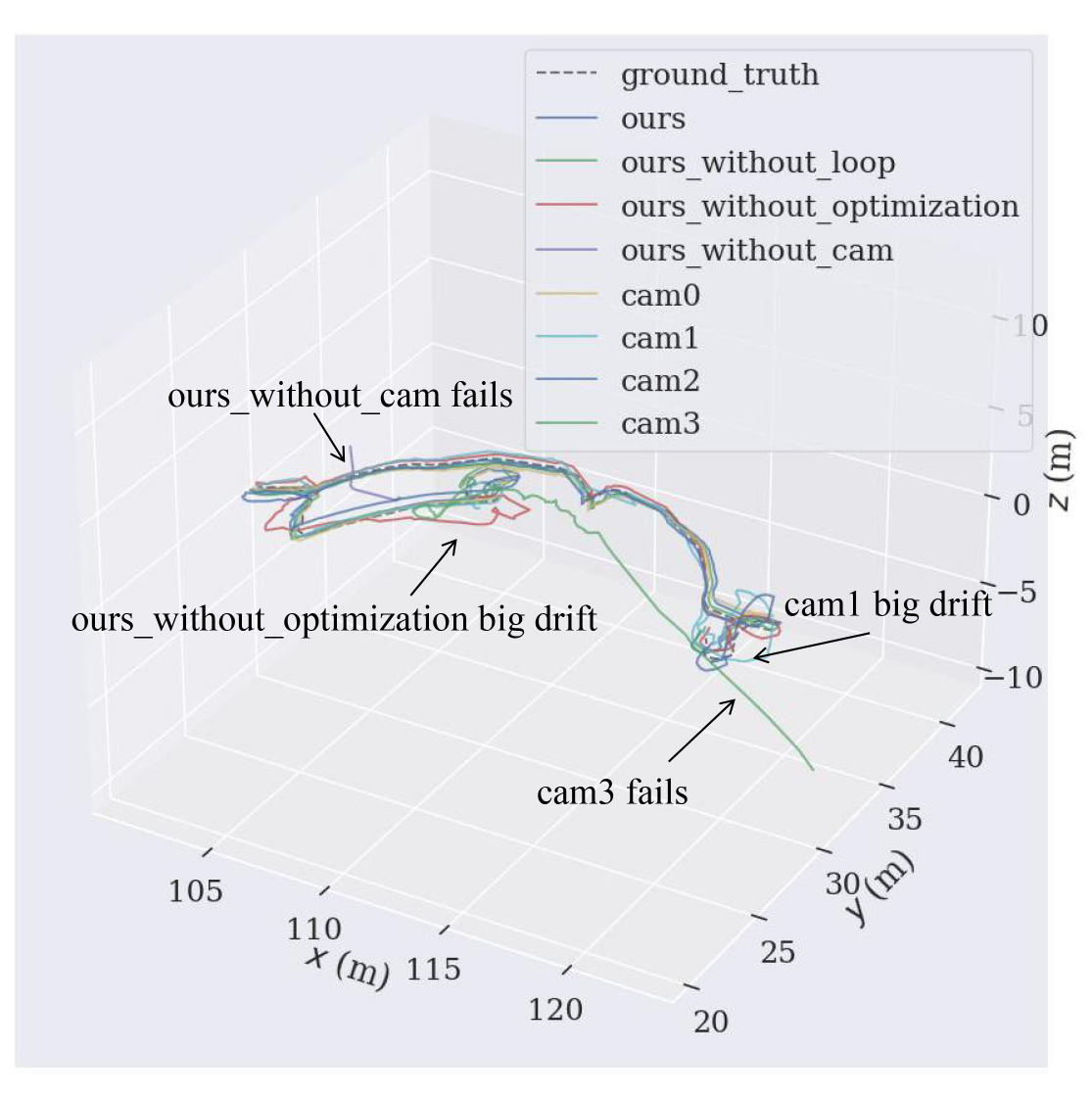} 
            \label{fig.Corridor-Lower-Gallery-2}
        }
        \caption{Pose trajectories estimated on the Construction Upper Level 3 and Corridor Lower Gallery 2 sequences of the Hilti'2022 Dataset.}
        \label{fig.hilti}
    \end{center}
\end{figure}

The results demonstrate the full configuration of the proposed method excels in all test sequences. 
It is noteworthy that in the M2DGR walk-01 test sequence (as shown in \Reffig{fig.walk-figure}), dynamic vehicles, moving pedestrians, and dense vegetation pose significant challenges to visual feature tracking. 
When relying solely on the mid-right camera, the odometry system experiences complete failure. 
This demonstrates that through multi-camera data fusion, the proposed system can leverage the complementary nature of different views to compensate for individual camera tracking failures, thereby significantly enhancing overall robustness.

Furthermore, the proposed extrinsic compensation algorithm performs better than without the compensation.
Taking the Newer College Stairs sequence as an example, because the degradation of LiDAR point cloud features increases the need for accurate triangulation results of visual features to provide accurate depth, the accuracy significantly deteriorates due to the offset between the multi-camera center and the panoramic model center. 
After implementing our compensation mechanism, the RMSE is reduced by 40.6$\%$, fully validating the effectiveness of the proposed method.

Due to the lack of sufficient LiDAR planar constraints in the Hilti'2022 test scenes, relying solely on LiDAR makes it difficult to achieve accurate and robust pose estimation, particularly in construction sites (Construction Upper Level 3) and stairwell environments (Corridor Lower Gallery 2). 
Additionally, strong outdoor exposure and extremely dark conditions in stairwells pose significant challenges for vision-based methods. 
In environments with insufficient feature textures, a single camera alone often fails to provide enough constraints to compensate for drift caused by LiDAR failure, and no single camera successfully passed all tests (as shown in \Reffig{fig.hilti}).
By fusing information from multiple cameras, we not only provide sufficient constraints for accurate pose estimation but also ensure robustness in scenarios where one camera lacks feature information by leveraging data from other cameras, preventing system failure. 
In cases where LiDAR fails and cannot provide depth information for visual feature points, our proposed extrinsic joint optimization method enables more accurate depth estimation for visual features, thereby improving pose estimation accuracy.

\subsection{Run Time Analysis}

\begin{table}[ht]
    \begin{center}
        \caption{The average runtime per image frame. (Unit: Millisecond)}
        \label{tab.runtime}
        \resizebox{\linewidth}{!}{
        \begin{tabular}{ccc}
            \hline\\[-2.9mm]\hline
            Dataset & one-camera & four-camera(ours) \\
            \hline
            Newer College Dataset & 47.927494 & 111.345361 \\
            \hline
            M2DGR Dataset & 45.911853 & 128.731194 \\
            \hline
            Hilti'2022 Dataset & 42.828289 & 54.051696 \\
            \hline
            Average & 45.555879 & 98.042750 \\
            \hline\\[-2.9mm]\hline
        \end{tabular}
        }
    \end{center}
\end{table}

We compare the average processing time between single-camera and four-camera configurations on three datasets: Newer College, M2DGR, and Hilti'2022.
As shown in \Reftab{tab.runtime}, the average processing time of the four-camera system is only 2.15 times that of the single-camera system. 
This result demonstrates that although the number of cameras increased fourfold, the proposed panoramic feature model and its efficient multi-camera feature fusion mechanism enable multi-view perception with only 2.15 times the computational overhead, effectively validating the superior computational efficiency of our approach.

\section{CONCLUSIONS}

In this paper, we propose Multi-LVI-SAM, a tightly coupled multi-sensor fusion framework that integrates multiple fisheye cameras, LiDAR, and IMU data to achieve robust and high-precision localization. 
To address the challenges of multi-camera fusion, we introduced a panoramic visual feature model, which unifies observations from multiple fisheye cameras into a globally consistent representation. 
Additionally, we propose an extrinsic compensation method to resolve triangulation inconsistencies caused by misalignment between individual camera frames and the panoramic model.
Through experimental evaluation on datasets in different platforms and environments, the proposed method shows higher accuracy and robustness comparing with existing methods.

The future work will focus on how to take advantage of the overlapping observation areas of neighboring cameras and the improvement of the robustness of the system in dark environments. 

\bibliographystyle{IEEEtran}
\bibliography{IEEEabrv,ref}
\end{document}